\documentclass[10pt,twocolumn,letterpaper]{article}

\usepackage{iccv}

\usepackage[accsupp]{axessibility}  
\usepackage{color}
\usepackage{times}
\usepackage{epsfig}
\usepackage{graphicx}
\usepackage{amsmath}
\usepackage{amssymb}
\usepackage{subfiles}
\usepackage{booktabs}
\usepackage{pifont}
\usepackage{multirow}
\usepackage{tabularx}

\usepackage{amsmath,amsfonts,bm}

\def\eqref#1{equation~\ref{#1}}

\def\1{\bm{1}}

\DeclareMathAlphabet{\mathsfit}{\encodingdefault}{\sfdefault}{m}{sl}
\SetMathAlphabet{\mathsfit}{bold}{\encodingdefault}{\sfdefault}{bx}{n}

\DeclareMathOperator*{\argmin}{arg\,min}

\usepackage{array}
\usepackage{multirow}
\newcolumntype{S}{>{\centering\arraybackslash}m{0.9cm}}
\newcolumntype{M}{>{\centering\arraybackslash}m{1.2cm}}
\newcolumntype{L}{>{\centering\arraybackslash}m{1.4cm}}
\usepackage[table]{xcolor}
\definecolor{mygray}{gray}{.95}
\definecolor{mylightergray}{gray}{.99}
\definecolor{mygreen}{RGB}{10, 179, 33}
\usepackage{algorithm}
\usepackage{algpseudocode}
\usepackage{setspace}

\usepackage[pagebackref=true,breaklinks=true,letterpaper=true,colorlinks,bookmarks=false]{hyperref}

\iccvfinalcopy 
\def\iccvPaperID{3910} 

\newcommand{\xrc}[1]{\textcolor{red}{#1}}

\usepackage{xparse}

\NewDocumentCommand{\INTERVALINNARDS}{ m m }{
    #1 {,} #2
}
\NewDocumentCommand{\interval}{ s m >{\SplitArgument{1}{,}}m m o }{
    \IfBooleanTF{#1}{
        \left#2 \INTERVALINNARDS #3 \right#4
    }{
        \IfValueTF{#5}{
            #5{#2} \INTERVALINNARDS #3 #5{#4}
        }{
            #2 \INTERVALINNARDS #3 #4
        }
    }
}

\usepackage{gensymb}

\definecolor{mygray}{gray}{.95}

\ificcvfinal\fi
\begin{document}

\title{An Empirical Study of the Collapsing Problem in Semi-Supervised \\  $2$D Human Pose Estimation}



\author{Rongchang Xie$^{1}$, ~Chunyu Wang$^{2}$, ~Wenjun Zeng$^{2}$, ~Yizhou Wang$^{3}$\\
	\normalsize $^{1}$Center for Data Science, Peking University ~~~~~~~~~~
	 $^{2}$Microsoft Research Asia 	\\
	\normalsize $^{3}$ Center on Frontiers of Computing Studies, CS Dept., Peking University\\
	{\tt\small \{rongchangxie, yizhou.wang\}@pku.edu.cn, \{chnuwa, wezeng\}@microsoft.com}
}

\maketitle
\ificcvfinal\thispagestyle{empty}\fi

\begin{abstract}
Most semi-supervised learning models are consistency-based, which leverage  unlabeled images by maximizing the similarity between different augmentations of an image. But when we apply them to human pose estimation that has extremely imbalanced class distribution, they often collapse and predict every pixel in unlabeled images as background. 
We find this is because the decision boundary passes the high-density areas of the minor class so more and more pixels are gradually mis-classified as background. In this work, we present a surprisingly simple approach to drive the model to learn in the correct direction. For each image, it composes a pair of easy-hard augmentations and uses the more accurate predictions on the easy image to teach the network to learn pose information of the hard one. The accuracy superiority of teaching signals allows the network to be ``monotonically'' improved which effectively avoids collapsing. We apply our method to the state-of-the-art pose estimators and it further improves their performance on three public datasets. The source code and pretrained models have been
released at \href{https://github.com/xierc/Semi\_Human\_Pose}{https://github.com/xierc/Semi\_Human\_Pose}.
\end{abstract}


\section{Introduction}
\label{section:introduction}

2D human pose estimation has many practical applications such as $3$D pose modeling \cite{zhang2020fusing,tu2020voxelpose,ma2021context} and action recognition \cite{wang2013approach,wang2016mining}. 
The early works in deep learning try to regress joint coordinates from images directly \cite{toshev2014deeppose,carreira2016human}. But most recent ones adopt the heatmap-based framework \cite{tompson2014joint,wei2016convolutional, newell2016stacked,sun2019hrnet,simplebaselines} because it provides better supervision. But there is a more important but less explored problem of learning robust models that perform well on unseen wild images. One solution is to fit the ``whole'' world by infinitely increasing training images. The other is to transfer pre-trained models to new domains by unsupervised finetuning. The common basis behind the two approaches is Semi-Supervised Learning (SSL)--- \emph{how to leverage unlabeled images to obtain a generalizable model? }

\begin{figure}[!tbp]
	\centering
	\includegraphics[width=1\linewidth]{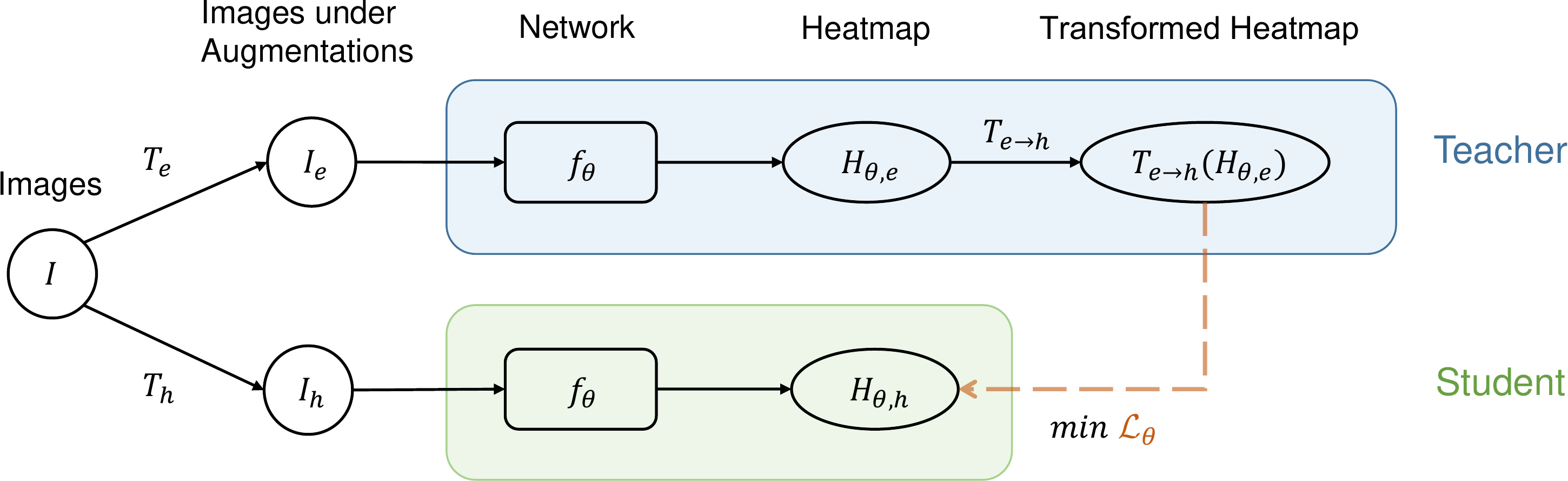}
	\caption{Our approach to avoid ``collapsing'' in semi-supervised human pose estimation. For each unlabeled image, we compose an easy and hard image pair $\mathbf{I}_e$ and $\mathbf{I}_h$ using two augmentation methods $T_e$ and $T_h$, respectively, and feed them to the network $f_{\theta}$. We use the heatmaps $T_{e\rightarrow h}{(\mathbf{H}_{\theta,e})}$ on the easy image to teach the network to learn about the hard image. $T_{e\rightarrow h}$ maps the two heatmaps of the augmented images. $\mathcal{L}_{\theta}$ represents the consistency loss. The accuracy superiority allows the network to be ``monotonically'' improved which avoids collapsing.}
	\label{fig:framework}
\end{figure}

The previous SSL works have primarily focused on the classification task. In general, there are two strategies to explore unlabeled images.
The first is Pseudo labeling \cite{radosavovic2018data,xie2020self} which first learns an initial model on only labeled images in a supervised way. Then, for each unlabeled image, it applies the initial model to obtain hard or soft pseudo labels representing its category. Finally, it learns the ultimate model on the combined dataset of labeled and pseudo-labeled images. However, the performance of the method is largely limited by that of the initial model which is learned only on the labeled images and fixed thereafter.

The second class of methods \cite{berthelot2019mixmatch,laine2017temporal,sajjadi2016ts,sohn2020fixmatch,tarvainen2017mean} learn about unlabeled images by requiring the network to have similar predictions for different augmentations of the same image. They are better than the pseudo labeling methods because the accuracy is not limited by the fixed labeling network. However, when we apply them to $2$D  pose estimation, we find that all of them encounter the collapsing problem meaning that, within few training iterations, the models begin to predict every pixel in unlabeled images as background. As a result, the prediction accuracy becomes even worse than the initial supervised model.

The collapsing problem is not identified as a serious issue in previous works because most of them were only evaluated on the well-balanced classification task. But we find it is vital for tasks with severe class imbalance such as human pose estimation, which has not received sufficient attention. It occurs because when the network makes different predictions on the corresponding pixels, it lacks sufficient information to determine the correct optimization path. Blindly minimizing their discrepancy causes the decision boundary to be incorrectly formed due to imbalance and pass through the high-density area of the minor class as revealed in \cite{hyun2020class}.
It leads to the situation where a growing number of pixels are mis-classified as background.

\begin{figure}[!htbp]
	\centering
	\includegraphics[width=1\linewidth]{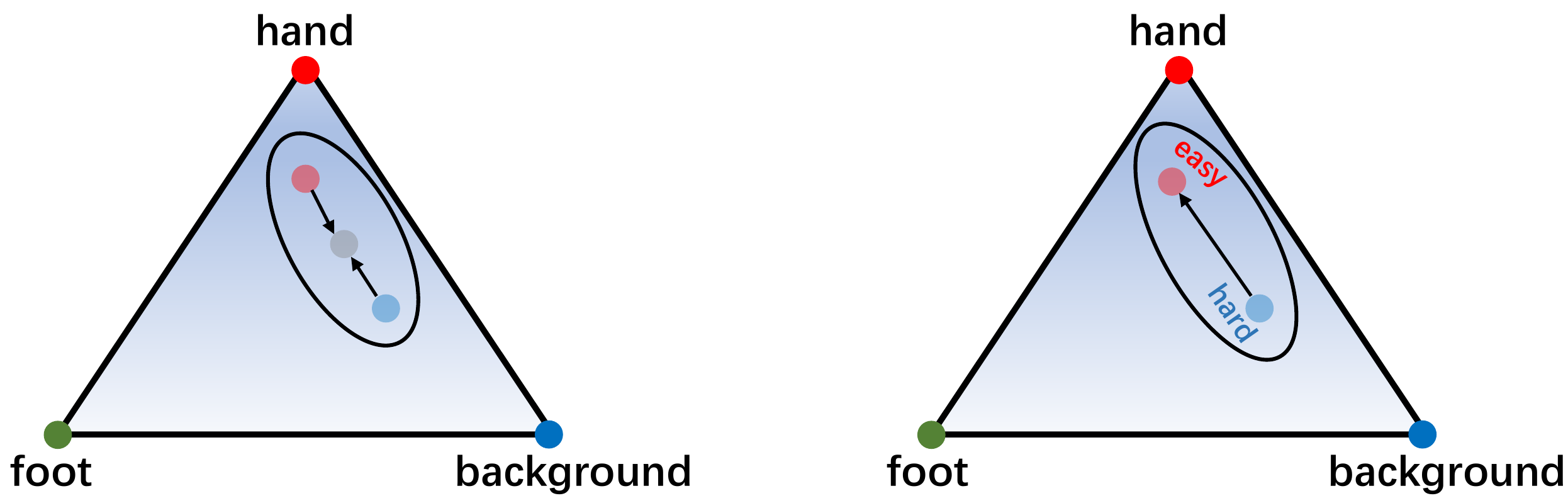}
	\caption{\textbf{Left:} the standard consistency-based method minimizes the distance between the predictions of the two augmentations (red and blue points). Since many pixels have low response (close to background), few high response pixels (\eg, the red point) tend to be gradually pulled to the background class. \textbf{Right:} In our method, more accurate predictions of easy augmentation pull those on hard augmentation, which avoids high response samples being pulled to the background class.  }
	\label{fig:collapsing}
\end{figure}

In this work, a simple approach is presented to address the collapsing problem. We first introduce the concept of \emph{easy-hard} augmentation pair and, by definition, a network should obtain better \emph{average} accuracy on a certain dataset with easy augmentation than on the same dataset with hard augmentation. Then, for each unlabeled image, we compose an easy and a hard augmentation, feed them to the network and obtain two heatmap predictions. We use the accurate predictions on the easy augmentation to teach the network to learn about the corresponding hard augmentation (see Figure \ref{fig:framework}). However, the hard augmentation will not be used for teaching the network to learn about the easy augmentation, which avoids high response samples being pulled to 
background as illustrated in Figure \ref{fig:collapsing}. The relative accuracy superiority of the teaching signals allows the network to be ``monotonically'' improved which stabilizes the training and avoids collapsing.

Our approach is general and applies to most consistency-based SSL methods such as \cite{ke2019dual,laine2017temporal} for stopping collapsing. We empirically validate it on a simple baseline as well as on the state-of-the-art method \cite{ke2019dual} which jointly learns two models. Both methods collapse in their original setting and our \emph{easy-hard} augmentation strategy helps avoid the problem.
We extensively evaluate them on three public datasets of COCO \cite{lin2014microsoft}, MPII \cite{andriluka14cvpr} and H36M \cite{ionescu2014human3}. When the number of labeled images is small, our approach increases the mean Average Precision (AP) by about $13\%$   (from $31.5\%$ to $44.6\%$) compared to the supervised counterpart which only uses labeled data for training. As a comparison, the pseudo labeling methods of \cite{xie2020self} and \cite{radosavovic2018data} only get $37.2\%$ and $37.6\%$ mean AP, respectively. More importantly, when we apply our method to the best $2$D pose estimator and use all available labeled training images, it can further improve the performance by a decent margin by exploring unlabeled images. We also report results when our approach is used for semi-supervised pre-training and domain adaptation tasks. The versatile practical applications in various settings validate the values of this work.

\section{Related Work}
\label{section:relatedwork}
SSL has been well studied for the classification task. We discuss some works which use deep networks since our target is to address the collapsing problem confronted by deep learning methods. Please refer to other surveys such as \cite{van2020survey} for a more comprehensive review. Pseudo labeling \cite{lee2013pseudo,radosavovic2018data,xie2020self,yarowsky1995unsupervised} is commonly used in SSL. The basic idea is to first learn an initial model on labeled images and then apply it to unlabeled images to estimate pseudo labels. The images with confident pseudo labels are added to the labeled dataset. Finally, it trains a stronger classifier on the extended dataset in a supervised way. However, the performance is limited by that of the initial classifier which is learned on only few labels. Iterative training alleviates the problem but the classifier is updated only once after it processes the whole dataset which is inefficient for large datasets. Besides, the selection criterion for data to be added to the labeled set is ad hoc for different tasks.

Some SSL methods \cite{sajjadi2016ts,laine2017temporal,tarvainen2017mean,berthelot2019mixmatch,sohn2020fixmatch} are consistency-based. For example, the $\Pi$ model \cite{laine2017temporal} keeps history predictions on the dataset and requires current predictions to be consistent with them. The approach is shown to be more tolerant to incorrect labels but is inefficient when learning large datasets since history predictions change only once per epoch. Tarvainen \etal \cite{ tarvainen2017mean} present the \emph{mean-teacher} model in which the teacher is the moving average of the student which can be timely updated in every iteration.  But their performance is limited because the two models tend to converge to the same point and stop further exploration.  Some methods \cite{qiao2018deep,ke2019dual} learn two different models by minimizing their prediction discrepancy. To avoid the case where the two models converge to the same point, they either learn from different initializations \cite{ke2019dual} or add view difference constraints \cite{qiao2018deep}.
Besides, There are some works that avoid collapsing without negative sample in self-supervised learning \cite{grill2020byol,chen2021exploring,zbontar2021barlow}, but their objective functions and optimized variables are different from ours. The BYOL uses the Exponential Moving Average (EMA) strategy \cite{grill2020byol}, which does not prevent collapsing in our experiments. The SimSiam shows that the stop-gradient plays an essential role \cite{chen2021exploring} but using it alone without our easy-hard augmentation strategy also cannot avoid collapsing.

The above works were not been evaluated for pose estimation and we find they all encounter the collapsing problem when applied to the task. The contribution of this work lies in identifying and studying the collapsing problem and presenting a simple solution to avoid it such that the existing SSL methods can be used for pose estimation. In addition,  we will extend some representative works to the human pose estimation task and provide a rigorous evaluation of their performance. This has empirical values to the community. We will release our code and models hoping it can facilitate research along this direction.

\section{The Method}
\label{section:easyhard}

The task of $2$D pose estimation aims to detect locations of $K$ body joints in an image $\mathbf{I}$. Since \cite{tompson2014joint}, nearly all methods transform the problem to estimating $K$ Gaussian heatmaps $\mathbf{H}$ where each heatmap encodes the probability of a joint at a location in $\mathbf{I}$. For inference, each joint can be estimated to be at the location with the largest value in the corresponding heatmap. Denote the labeled and unlabeled training sets as $\mathcal{L}=\{(\mathbf{I}^l, \mathbf{H}^{l})\}_{l=1}^{N}$ and    $\mathcal{U}=\{\mathbf{I}^u\}_{u=1}^{M}$, respectively. For supervised training of the pose estimation network $f$, we minimize the MSE loss between the estimated and ground-truth heatmaps:

\begin{equation}
L_s = \mathop{\mathbb{E}_{\mathbf{I}\in{\mathcal{L}}}  ~||f(\mathbf{I}_{\eta},\theta)-\mathbf{H}_{\eta}||^2},
\label{eq:supervised_loss}
\end{equation}
where $\mathbf{I}_{\eta}=T(\mathbf{I}, \eta)$ represents an augmentation of $\mathbf{I}$ and $\eta$ represents augmentation parameter. $\mathbf{H}_{\eta}=T(\mathbf{H}, \eta)$ represents the corresponding heatmap and $\theta$ represents the network parameters.

\subsection{Unsupervised Learning via Consistency}
The network $f$ also learns about unlabeled images via consistency loss. For each unlabeled image $\mathbf{I}$, it composes two augmentations $\mathbf{I}_{\eta}$ and  $\mathbf{I}_{\eta'}$ and minimizes the MSE loss between the heatmap predictions:
\begin{equation}
L_u = \mathop{\mathbb{E}_{\mathbf{I}\in{\mathcal{U}}}  ~||f(\mathbf{I}_{\eta},\theta)-f(\mathbf{I}_{\eta'},\theta')||^2}.
\label{eq:cons}
\end{equation}
The network parameters $\theta$ and $\theta'$ can be either identical or different. For example, in \cite{tarvainen2017mean}, $\theta'$ is the exponential
moving average (EMA) of $\theta$. We will evaluate both choices in our experiments. \emph{It is worth noting that both $\theta$ and $\theta'$ are changing during training. In contrast, the teacher network of the pseudo labeling methods is fixed so it does not suffer from collapsing.} The parameters $\eta$ and $\eta'$ are usually randomly sampled at each training step. It is worth noting that $\eta$ and $\eta'$ are usually sampled from the \emph{same} distribution without discrimination \cite{berthelot2019mixmatch,laine2017temporal,tarvainen2017mean}.

\begin{figure}[!htbp]
	\centering
	\includegraphics[width=1\linewidth]{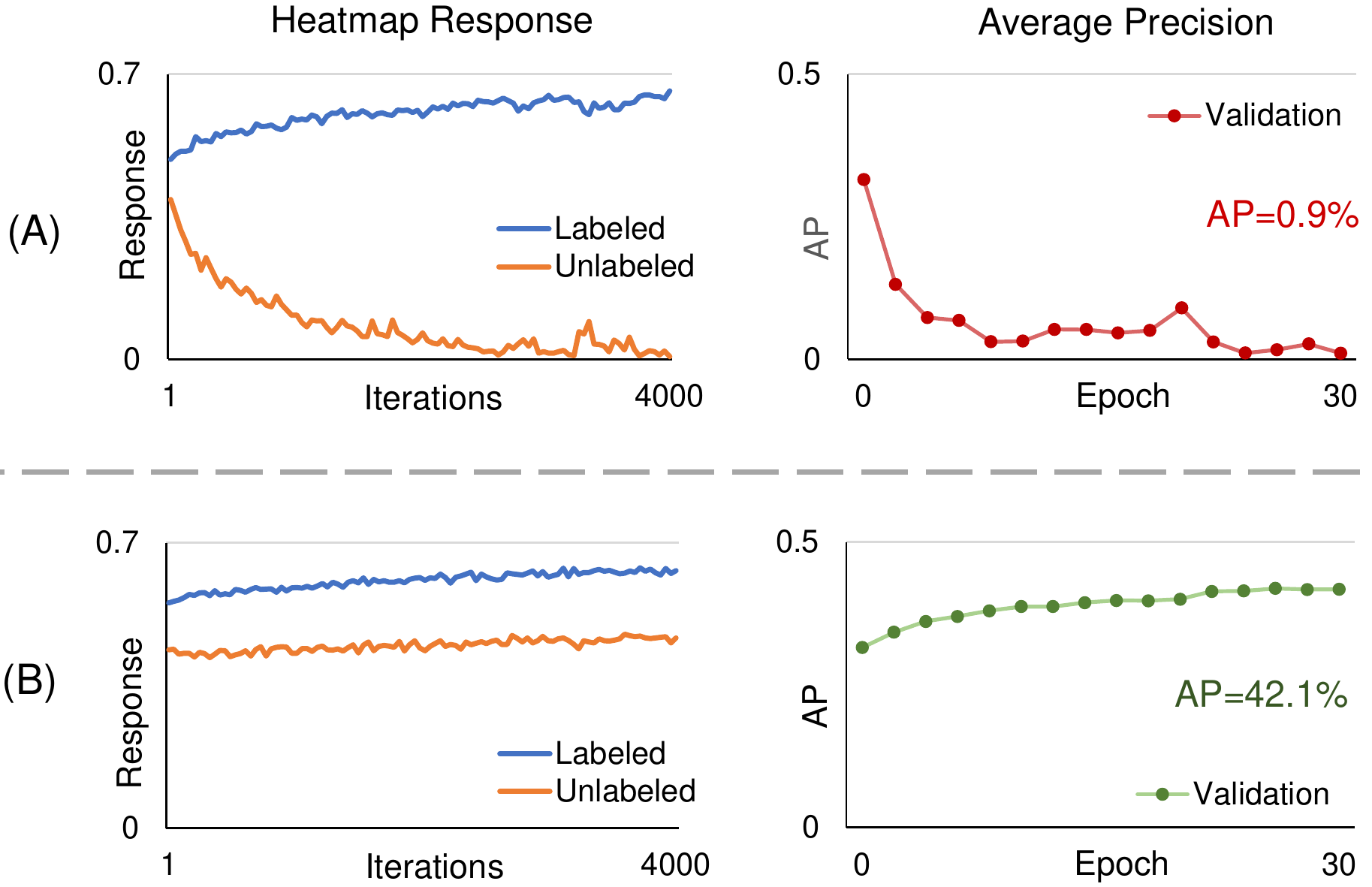}
	\caption{\textbf{Top:} results of the standard consistency-based method. Average heatmap response increases steadily for labeled images which is as expected. But for unlabeled images, it decreases to zero which suggests that collapsing occurs. The estimation accuracy on the validation dataset also decreases to $0.9\%$. \textbf{Below:} the results of our approach.}
	\label{fig:easy}
\end{figure}

\subsection{The Collapsing Problem}
We try to train a model by adding the two loss functions: $L=L_s+\lambda L_u$ with $\lambda=1$. Each batch of training data consists of equal number of images from $\mathcal{L}$ and $\mathcal{U}$. We use affine augmentation \cite{sun2019hrnet,simplebaselines} for $\eta$ and $\eta'$. We use identical weights for $\theta$ and $\theta'$ and use $1$K labels. Within a few iterations of training, the network begins to predict all pixels of unlabeled images as background as shown in Figure \ref{fig:easy} (top). The maximum value in a heatmap is used to represent its heatmap response and we find the average response on labeled images increases steadily which is as expected. However, the average response on unlabeled images decreases significantly and the accuracy on validation images is very low. Decreasing $\lambda$ does not solve the problem. It only slows down the collapsing process. So we set $\lambda = 1$ for the rest of our experiments. Some one may think it is over-fitting to the small labeled dataset. However, increasing labels to $118K$ does not fully solve the problem. The response on unlabeled images still gradually decreases. The accuracy is higher than the case with $1$K labels but it is still worse than the initial superivsed model. We also tried to use strong augmentation methods such as Rand Augmentation \cite{cubuk2020randaugment} to labeled images or unlabeled images but none of them can fully address the collapsing problem.

\begin{figure}[!htbp]
	\centering
	\includegraphics[width=1\linewidth]{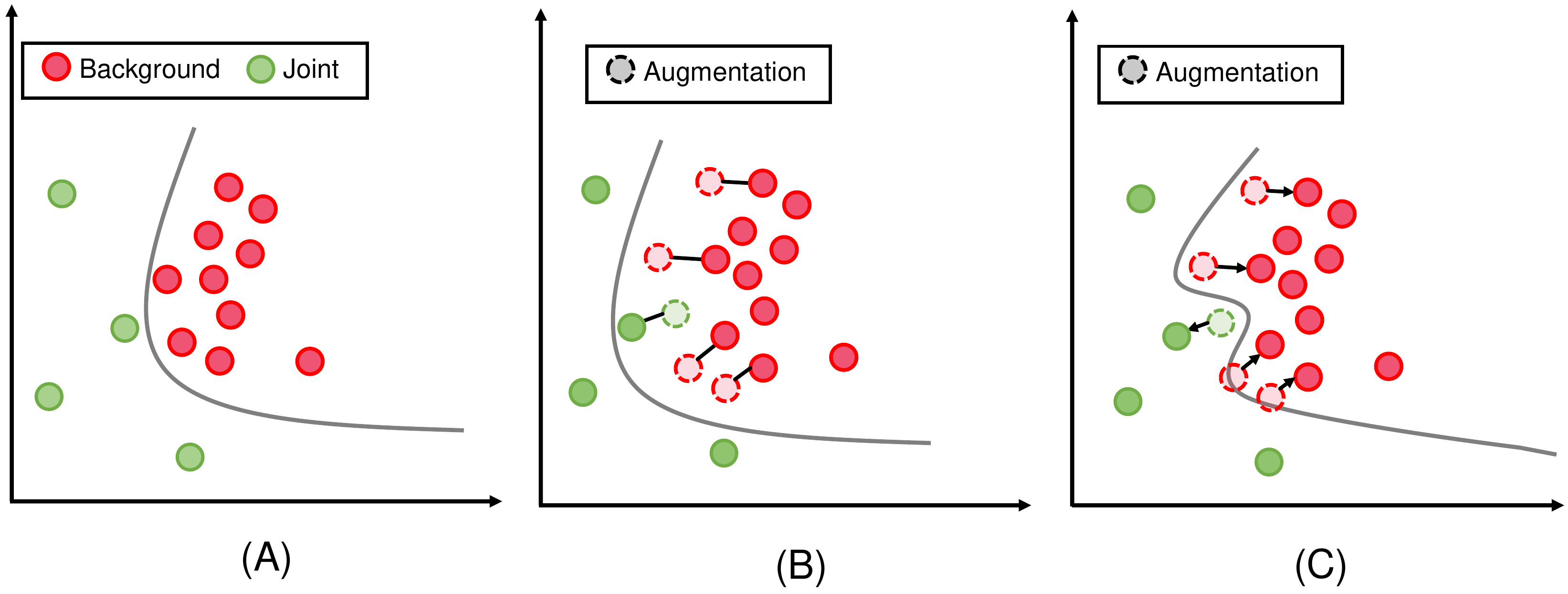}
	\caption{\textbf{(A)} the decision boundary before SSL. \textbf{(B)} the na\"ive consistency regularization moves data and their augmentations (dashed circles) to their middle points. As a result, more data will be close to the decision boundary which pushes the decision boundary to pass through the areas of minor class that is sparse globally. \textbf{(C)} differently, our approach drives the less accurate predictions, which are close to the decision boundary, to the direction of more accurate predictions. In this case, the decision boundary is less likely to be incorrectly formed.}
	\label{fig:intuition}
\end{figure}

Collapsing occurs because the consistency regularization requires the model to satisfy the smoothness assumption \cite{ssl_book,van2020survey} where an image and its augmentation should have similar predictions. Thereby, the decision boundary would be pushed to low-density region. In fact,
due to the imbalance in data, decision boundary often skews into the areas of minor class which is sparse globally as shown in Figure \ref{fig:intuition}. This is also observed in \cite{hyun2020class}. As a result, a growing number of pixels are mis-classified as background.

\subsection{Avoid Collapsing}
The na\"ive implementation of the consistency regularization draws two samples to their middle point so more data are becoming closer to the decision boundary (see Figure \ref{fig:intuition}.B). As a result, the decision boundary is pushed away from the high density areas of the dominant class and may skew into the areas of minor class. In contrast, our approach drives the less accurate predictions which are close to the decision boundary to the direction of more accurate predictions. In this case, the decision boundary is less likely to skew into the areas of minor class. 

To achieve the goal, we present a paired easy-hard image augmentation strategy. For an unlabeled image $\mathbf{I}$, it obtains two augmented images $\mathbf{I}_e$ and $\mathbf{I}_h$ by applying an easy and hard augmentation $T_e$ and $T_h$, respectively:
\begin{equation}
    \mathbf{I}_e = T_e(\mathbf{I})=T(\mathbf{I},\eta_e) ~~ \text{and} ~~
     \mathbf{I}_h = T_h(\mathbf{I})=T(\mathbf{I},\eta_h).
\end{equation}

Where $T_e$ is regarded as an easier augmentation method than $T_h$ only when the network obtains better \emph{average accuracy} on a dataset under perturbation $T_e$ than under $T_h$. We feed the two augmented images to the network and let the predictions of $\mathbf{I}_e$ to teach the predictions of $\mathbf{I}_h$:

\begin{equation}
L_{e,h} = \mathop{\mathbb{E}_{\mathbf{I}\in{\mathcal{U}}}  ~||f(\mathbf{I}_e,\theta)-f(\mathbf{I}_h,\theta)||^2}.
\label{eq:easyhard}
\end{equation}

For the sake of simplicity, we call $f(\mathbf{I}_e,\theta')$ and $f(\mathbf{I}_h,\theta)$ as teacher and student signals, respectively. \emph{Note that the gradients are propagated through only the student path.} This is the key to avoid collapsing. This can be done by calling the detach operator on the teacher signals before computing the loss. Removing the detach operator leads to collapsing regardless of augmentations.

\section{Implementation Details}
\begin{figure}[!htbp]
	\centering
	\includegraphics[width=1\linewidth]{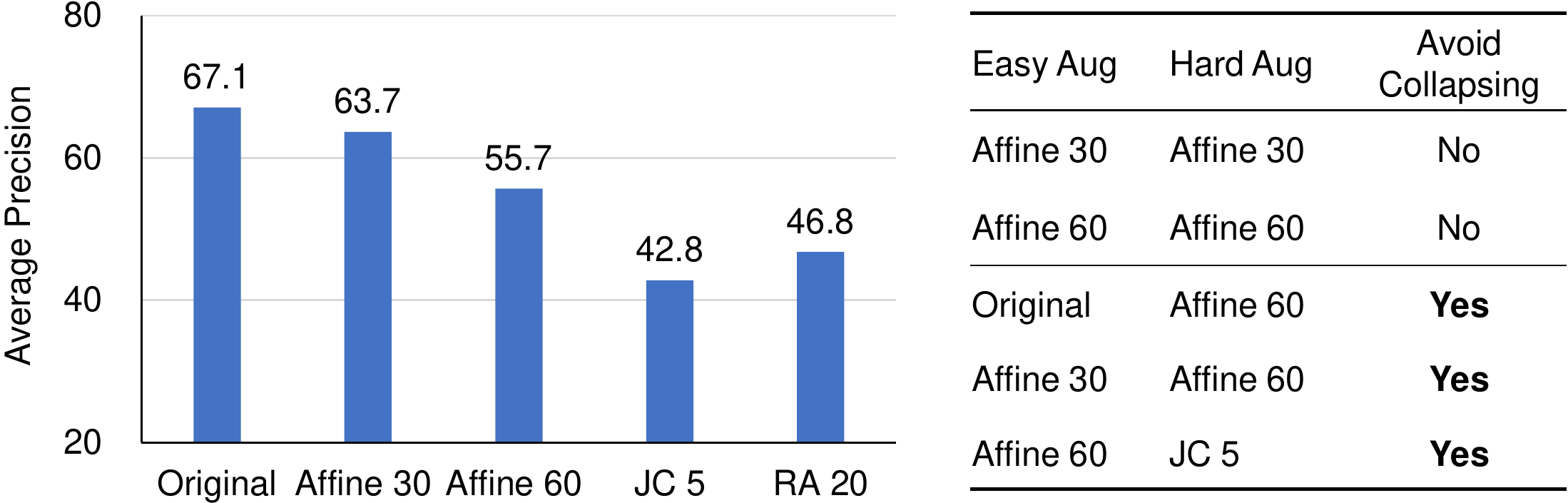}
	\caption{\textbf{Left:} Average precision of a network on a dataset under different augmentation. \textbf{Right:} Some example easy-hard augmentation pairs and their effects on avoiding collapsing. ``JC 5'' represents Joint Cutout Augmentation on five joints (a novel hard augmentation method we introduce in section \ref{section:implementation}). ``RA 20'' represents Random Augmentation \cite{cubuk2020randaugment}. }
	\label{fig:augmentation_strength}
\end{figure}

\label{section:dualnetwork}
\subsection{Easy-Hard Augmentation}
\label{section:implementation}
\paragraph{Affine Transformation} is commonly used in $2$D pose estimation which randomly scales and rotates an image. Affine transformation changes keypoint locations for  pose estimation which are equivariant to the transformation \cite{zhang2018unsupervised,thewlis2017unsupervised}.
Let $T(\cdot)$ be an affine transformation and $f(\cdot)$ be the network to estimate heatmaps from images. Then the loss function can be computed as:
\begin{equation}
L = \mathop{\mathbb{E}_{\mathbf{I}\in{\mathcal{U}}}  \| f( T(\mathbf{I}) ) - T( f(\mathbf{I}) )\|^2}.
\label{eq:eqv_reg}
\end{equation}
It can be extended to map the heatmaps of the same image under two different affine augmentations which allows us to compute the consistency loss.

We find that a pose estimator achieves very different performances on the same dataset if we apply affine transformation of different \emph{strengths} to perturb the testing images. Figure \ref{fig:augmentation_strength} shows some typical results. For example, when we randomly sample rotation angles from $[-30^{\circ}, 30^{\circ}]$ and scale factors from $[0.75, 1.25]$ (denoted as ``Affine 30'') for affine transformation to perturb testing images, the Average Precision (AP) on the dataset is about $63.7\%$. But when we sample from a larger range of $[-60^{\circ}, 60^{\circ}]$ and $[0.5, 1.5]$, respectively, AP decreases notably to $55.7\%$.

The above finding motivates us that we can compose easy-hard augmentation pairs by adapting the ranges of rotation and scaling. Figure \ref{fig:augmentation_strength} shows some easy-hard augmentation choices that are able to prevent the model from collapsing.
It is worth noting that ``Affine 60'' can be regarded as a hard augmentation compared to ``Affine 30'', but it can also be regarded as an easy augmentation when compared to a stronger method ``JC 5'' which we will introduce in the next section. \emph{It suggests that it is the gap between the two methods that matters}.

Note that the augmentation strategies generalize well across datasets, which means that we need not to repeat the experimentation. In our experiments, the augmentations are determined based on 1K images sampled from COCO,
and are applied to the rest datasets.

\begin{figure}[!htbp]
	\centering
	\includegraphics[width=1.0\linewidth]{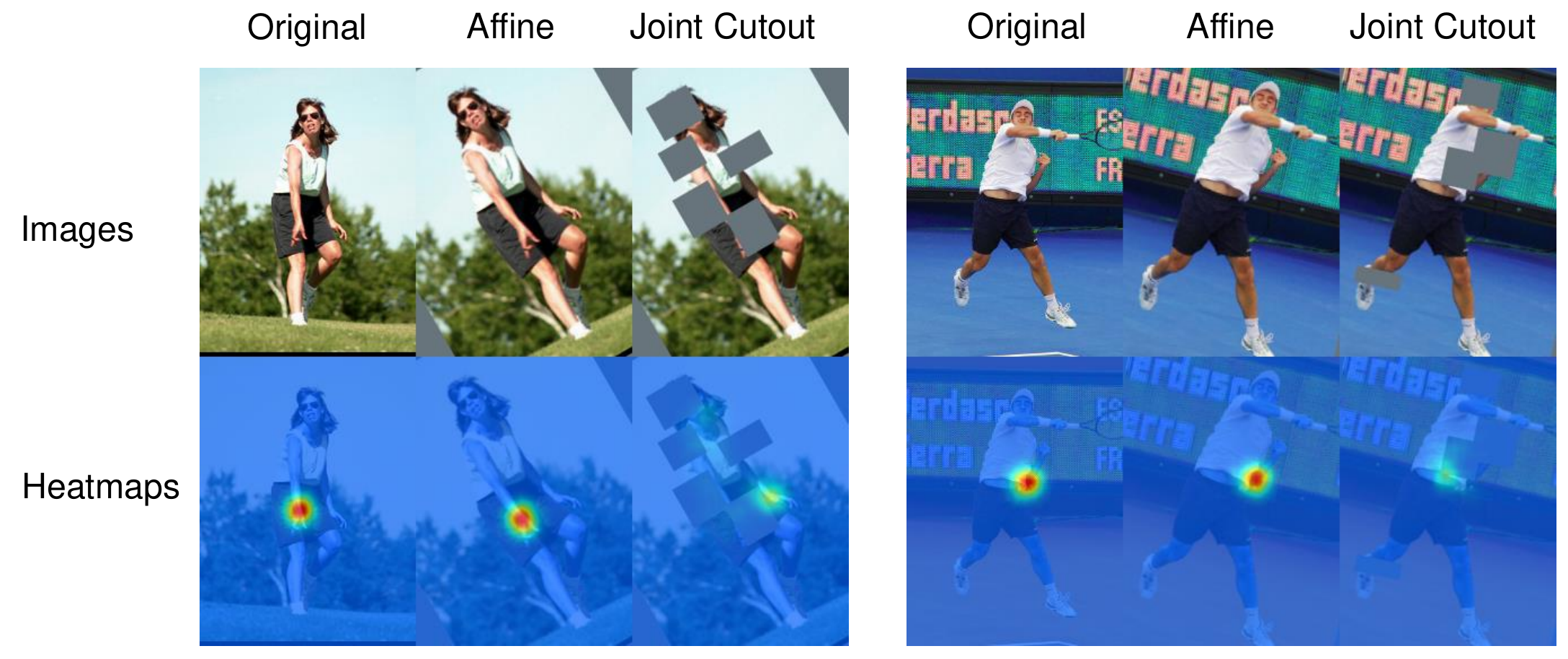}
	\caption{Effect of Joint Cutout. In each example, we show original, affine transform and Joint Cutout images and heatmaps. We can see that Joint Cutout is more effective in fooling the network. In the left example, since the head is occluded, the model has difficulty in discriminating left and right joints, which drives the model to learn more discriminative features.  }
	\label{fig:jointcutout}
\end{figure}

\paragraph{Joint Cutout}
Although affine-based augmentation already avoids collapsing, we find using harder augmentation for $T_h$ improves the accuracy ($T_e$ still uses easy affine augmentation). 
Inspired by cutout \cite{devries2017improved} and keypoint masking \cite{ke2018multi}, we introduce a new method \emph{Joint Cutout} to simulate occlusion. For each image (with easy augmentation applied), we first estimate coarse locations of keypoints using the model we are trying to train. Then we randomly sample a number of detected keypoints and mask their surrounding regions as illustrated in Figure \ref{fig:jointcutout}. To avoid over-fitting to the masks, the center locations and sizes of masking regions are randomly perturbed. The method improves accuracy by a notable margin especially when the number of labeled images is small.

\subsection{Learning Dual Networks}
\label{section:dual}
The previous SSL methods \cite{tarvainen2017mean,laine2017temporal,sohn2020fixmatch,berthelot2019mixmatch} often learn a single network where the teacher's parameters are either the same as the student's or its exponential moving average. So the teacher and student networks are coupled which limits their performance \cite{ke2019dual}. The recent method \cite{ke2019dual} learns two independent networks to solve the problem. In this section, we briefly introduce how to apply easy-hard augmentation to it. For a training image $\mathbf{I}$, we generate an easy and a hard augmentation denoted as $\mathbf{I}_e$ and $\mathbf{I}_h$, respectively. Then we feed them to both networks $f_{\theta}$ and $f_{\xi}$ and obtain four stream heatmap predictions:
\begin{equation}
\label{eq:heatmap}
\begin{split}
       & \mathbf{H}_{\theta, e} = f(\mathbf{I}_e,\theta) \quad \text{and} \quad
    \mathbf{H}_{\theta, h} = f(\mathbf{I}_h,\theta) ,\\
   & \mathbf{H}_{\xi, e} = f(\mathbf{I}_e,\xi) \quad \text{and} \quad
    \mathbf{H}_{\xi, h} = f(\mathbf{I}_h,\xi) .
\end{split}
\end{equation}

We know that $\mathbf{H}_{\theta, e}$ and $\mathbf{H}_{\xi, h}$ are similar up to a known transformation $T_{e \rightarrow h}$. Similarly, $\mathbf{H}_{\theta, h}$ is also similar to $\mathbf{H}_{\xi, e}$ up to the same transformation. We train the networks by minimizing two consistency loss items:
\begin{equation}
\label{eq:loss}
\begin{split}
& \theta^{*} = \argmin_{\theta} {\| \mathbf{H}_{\theta, h} - T_{e \rightarrow h}(\mathbf{H}_{\xi, e})\|^2} , \\
& \xi^{*} = \argmin_{\xi} {\| \mathbf{H}_{\xi, h} - T_{e \rightarrow h}(\mathbf{H}_{\theta, e})\|^2} .\\
\end{split}
\end{equation}

We only pass the gradient back through the hard example to avoid collapsing. It means that one consistency loss item is used to optimize a single network at each time. 
Take the first formula in Eq. (\ref{eq:loss}) as an example, $\mathbf{H}_{\xi, e}$ estimated by $f_{\xi}$ is treated as a teacher to update $f_{\theta}$. In this case, we do not update $f_{\xi}$ because $\mathbf{H}_{\theta, h}$ is usually too noisy to be used as supervision. Subsequently, we update $f_{\xi}$ according to the second formula in Eq. (\ref{eq:loss}).
The two symmetrical loss items are combined so that the two networks can guide each other and be optimized together. The performances of the two networks are very close in the end and we report their average accuracy.  Note that in inference, the model has the same number of parameters and running speed as the supervised model.

\section{Baselines and Our Methods}

We first introduce several baselines by modifying some representative SSL classifiers for pose estimation including both Pseudo labeling methods and consistency-based ones, and numerically compare them to our approach.

\paragraph{PseudoPose}
It is modified from pseudo labeling methods \cite{lee2013pseudo,radosavovic2018data,xie2020self}. We first train a teacher model $f_t$ with labeled images. Then $f_t$ is fixed and we apply it to unlabeled images to obtain pseudo heatmaps. We train an ultimate model $f$ by minimizing the Mean squared error (MSE) loss on the combined set:
\begin{equation}
    L = \sum_{\mathbf{I} \in\mathcal{U}} \|  T_{e \rightarrow h}(f_{t}(\mathbf{I}_e)) -   f(\mathbf{I}_h)  \|^2 + \sum_{\mathbf{I} \in\mathcal{L}} \|  \mathbf{H}_e -   f(\mathbf{I}_e)  \|^2,
\label{eq:self}
\end{equation}
where $\mathbf{H}_e$ is the ground-truth heatmap. Note that we use the same augmentation methods as ours for fair comparison.

\paragraph{DataDistill \cite{radosavovic2018data}} It is also a pseudo labeling method. It differs from \textbf{PseudoPose} in that it \emph{sums} the heatmaps estimated for multiple different augmentations of an image, obtains the keypoint locations, and re-generates a pseudo heatmap with Gaussian shape for supervision.
 
\paragraph{Ours (Single)} It is a consistency-based method in which $\theta$ and $\theta'$ are identical. On labeled images, it performs supervised learning with the ground-truth heatmaps. For each unlabeled image, it minimizes the discrepancy between the two estimated heatmaps of the \emph{easy and hard augmented images}. It differs from \textbf{PseudoPose} in that $f_t$ is not fixed. In fact, it is $f$ which is learned in semi-supervised learning.

\paragraph{Ours (Dual)} The method is similar to ``Ours (Single)'' except that it learns dual networks as discussed in section \ref{section:dual}. We also apply our proposed ``easy-hard'' augmentation method to this approach to avoid collapsing.

\section{Experiment}
\subsection{Datasets, Metrics and Details}

\paragraph{COCO Keypoint \cite{lin2014microsoft}}
It has four subsets of \emph{TRAIN},
\emph{VAL}, \emph{TEST-DEV} and \emph{TEST-CHALLENGE}. There are $123$K \emph{WILD} unlabeled images. To evaluate our method when different numbers of labels are used, we construct four training sets by randomly selecting $1$K, $5$K, $10$K and $20$K person instances from \emph{TRAIN}, respectively. The unlabeled set consists of the rest of images from \emph{TRAIN} unless specified. In some experiments, we use the whole \emph{TRAIN} as the labeled set and \emph{WILD} as the unlabeled one. We report the mean AP over $10$ OKS thresholds as the main metric following \cite{lin2014microsoft}. The input image size is $256 \times 192$.

\begin{table}[!htbp]
	\centering
	\caption{AP of different methods on COCO when different numbers of labels are used. The bottom section (grayed) evaluates augmentation methods. ``A'' represents Affine transformation.}
	\label{table:baselines}
	\begin{tabular}{llcccc}
		\toprule
		Methods     & Aug. & $1$K   & $5$K   & $10$K  & All               \\
		\hline
		Supervised \cite{simplebaselines} & A  & 31.5          & 46.4          & 51.1         & 67.1 \\
		\hline
		PseudoPose  & A  & 37.2          & 50.9          & 56.0             & ---                        \\
		DataDistill \cite{radosavovic2018data} & A & 37.6 & 51.6 & 56.6 & ---\\
		\hline
		Ours (Single)     & A & 38.5          & 50.5          & 55.4             & ---                        \\
		\textbf{Ours (Dual)} & \textbf{A}  & \textbf{41.5} & \textbf{54.8} & \textbf{58.7}  &  ---\\
		\toprule
	    \rowcolor{mygray}
		Ours (Single) & A+JC & {42.1} &{52.3} &{57.3} & --- \\
		\rowcolor{mygray}
		Ours (Dual) & A+RA & 43.7 & 55.4 & 59.3 & --- \\
		\rowcolor{mygray}
		Ours (Dual) & A+JC & 44.6 & 55.6 & 59.6 & --- \\
		\bottomrule                       
	\end{tabular}
\end{table}

\paragraph{MPII Dataset \cite{andriluka14cvpr}}
It has about $25$K images with $40$K annotated person instances. Since labels are not provided for the test set, we conduct ablation study on the validation set which consists of $3$K instances. We use the training set as the labeled set and the AI Challenger dataset \cite{wu2019large} as the unlabeled set, which has 210K
images with 370K person instances. The metric of PCKh@$0.5$ \cite{andriluka14cvpr} is reported. The size of the input image is $256 \times 256$ following \cite{andriluka14cvpr}.

\paragraph{H36M Dataset \cite{ionescu2014human3}}
We use subjects S1, S5, S6, S7 and S8 for training, and S9, S11 for testing.
The 2D pose accuracy is measured by Joint Detection Rate (JDR). And the Mean Per Joint Position Error (MPJPE) is used as the main metric in 3D pose estimation.

\paragraph{Implementation Details}
We use \emph{SimpleBaseline} \cite{simplebaselines} to estimate heatmaps and ResNet18 \cite{he2016deep} as its backbone. But our approach is general and can be applied to other pose estimators as shown in Table \ref{table:cocotest}.
On the validation set, we use the ground truth boxes and do not flip images for all methods. We train the models for $100$ epochs. We use Adam \cite{kingma2014adam} optimizer with the initial learning rate of
$1e^{-3}$. It drops to $1e^{-4}$ and $1e^{-5}$
at $70$ and $90$ epochs, respectively.

\begin{figure*}
	\centering
	\includegraphics[width=1\linewidth]{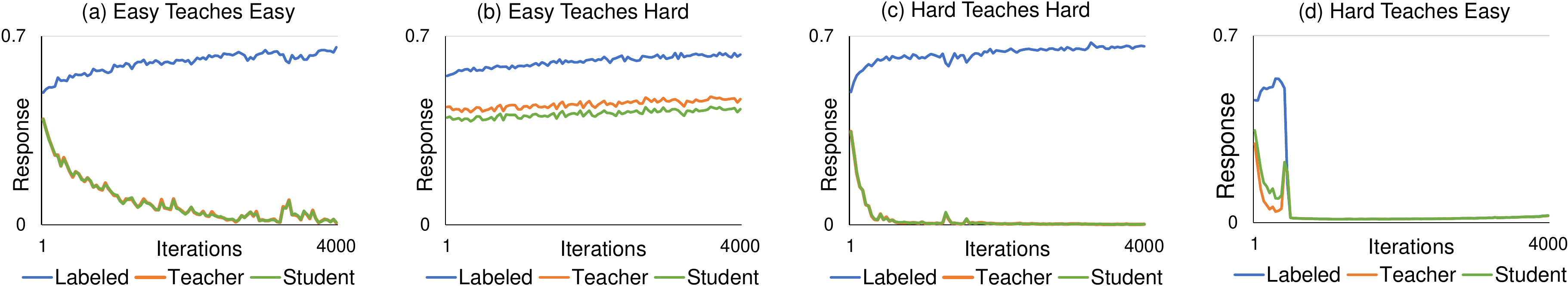}                                                                   
	\caption{Evolution of average heatmap around body joints of different augmentation strategies. The blue line represents the results on labeled images. The red and green lines represent the results of the teacher and student, respectively, on unlabeled images. }
	\label{fig:score_change_per_iteration}
\end{figure*}

\begin{table}[]
    \footnotesize   
	\centering
	\caption{The effects of using different network structures for the two models $f_{\theta}$ and $f_{\xi}$ on COCO. We report AP when different numbers of labels are used.}
	\label{table:diffmodels}
    \begin{tabular}{llccc}
    \toprule
    \multicolumn{1}{l}{Method}   & Networks of $f_{\theta}$ and $f_{\xi}$   & $1$K           & $5$K           & $10$K          \\
    \midrule
    Supervised \cite{simplebaselines} & ResNet18      & 31.5          & 46.4          & 51.1          \\
    Supervised \cite{simplebaselines} & HRNet w48  & 39.2          & 57.7          & 63.7          \\
    \hline
    \multirow{2}{*}{Ours (Dual)}        & ResNet18     & 41.5          & 54.6          & 58.6          \\
                                 & ResNet18     & 41.6          & 54.9          & 58.8          \\
    \hline
    \multirow{2}{*}{Ours (Dual)}        & HRNet w48     & 50.9  & 64.3  & 67.9  \\
                                 & HRNet w48 & 51.0         & 64.2          &  67.9    \\
    \hline
    \multirow{2}{*}{Ours (Dual)}        & ResNet18     & \textbf{48.7} & \textbf{59.4} & \textbf{62.5} \\
                                 & HRNet w48 & 50.9          & 62.8          & 66.8    \\

    \bottomrule 
    \end{tabular}
\end{table}

\subsection{Ablative Study}

\paragraph{Easy-Hard Augmentation}
We first study the relationship between augmentation methods and collapsing.
As shown in Figure \ref{fig:score_change_per_iteration} (a), when we use easy augmentations for both $T_e$ and $T_h$, the average response gradually decreases to zero for unlabeled images meaning collapsing occurs. This is because there is no accuracy gap between teacher and student signals. We also get degenerated results when we use hard augmentation for $T_e$ (see sub-figures c and d) for the same reason. In contrast, the training becomes normal when we use easy-hard augmentation strategy. In this case, the teacher and student models have sufficient gap. We have similar observation when we either learn a single model or dual models.

\paragraph{Baseline SSL Methods}
Table \ref{table:baselines} shows the results of different SSL methods. Supervised training with a small number of labels gets worst results which validates the values of unlabeled images. The gap is larger when there are fewer labels. DataDistill \cite{radosavovic2018data} achieves slightly better accuracy than PseudoPose since it ensembles multiple network output  to obtain more reliable pseudo labels. The proposed consistency-based method "Ours (Dual)" get better results than the pseudo labeling methods.

We also study the impact of augmentation methods for $T_h$. We can see from Table \ref{table:baselines} (bottom) that applying harder augmentation methods such as RandAug (RA) and Joint Cutout (JC) notably improves the results especially when the number of labeled images is small. In particular, Joint Cutout achieves consistently better mean AP scores than  RandAug. It is known that the most common mistake in pose estimation is the confusion between left and right joints. As shown in Figure \ref{fig:jointcutout}, Joint Cutout is effective in increasing the level of confusion and drives the models to learn more discriminative features. \emph{We use Joint Cutout as the default augmentation for the rest of the paper}. It is worth noting that applying hard augmentation to \cite{simplebaselines} in supervised training actually decreases AP when there are $1$K labels and slightly increases AP from $51.1\%$ to $52.1\%$ when there are $10$K labels.  

\begin{table}[!htbp]
	\small
	\renewcommand\tabcolsep{3.5pt}
	\caption{Results on the COCO \emph{VAL} set when all images from the \emph{TRAIN} set are used as the labeled set and all images from the \emph{WILD} set are used as the unlabeled set.}
	\label{table:largescale}
	\begin{tabular}{lllccc}
		\toprule
		Method & Network & AP & Ap .5 & AR & AR .5 \\
		\hline
		Supervised \cite{simplebaselines} & ResNet50  & 70.9        & 91.4            & 74.2        & 92.3  \\
		Ours & ResNet50      & \textbf{73.9 \xrc{\footnotesize ($\uparrow$3.0)}} & 92.5 & 77.0 & 93.5 \\
		\hline
		Supervised \cite{simplebaselines} & ResNet101     & 72.5        & 92.5           & 75.6        & 93.1  \\
		Ours & ResNet101     & \textbf{75.3 \xrc{\footnotesize ($\uparrow$2.8)}}       & 93.6           & 78.2        & 94.1     \\
		\hline
		Supervised \cite{simplebaselines} & ResNet152 &      73.2        & 92.5           & 76.3        & 93.2      \\
		Ours & ResNet152     & \textbf{75.5 \xrc{\footnotesize ($\uparrow$2.3)}}      & 93.6            & 78.5        & 94.3     \\
		\hline
		Supervised \cite{sun2019hrnet} & HRNetW48      & 77.2        & 93.5           & 79.9        & 94.1    \\
		Ours & HRNetW48    & \textbf{79.2 \xrc{\footnotesize ($\uparrow$2.0)}}       & 94.6 & 81.7 & 95.1  \\
		\bottomrule         
	\end{tabular}
\end{table}

\paragraph{Network Structures}
We evaluate the effect of using different networks in Table \ref{table:diffmodels}. We can see that when we use ResNet18 and HRNet, the performance of ResNet18 improves by a large margin ($41.5\%$ vs. $48.7\%$) compared to the case of using ResNet18 for both networks. This is mainly because HRNet can provide more accurate supervision for ResNet18 which notably boosts its performance. The results suggest that, even when our target is to learn a lightweight model for fast inference, we can still learn it together with a large model  which will notably improve the accuracy of the lightweight model.

\subsection{Failed Attempts}
\begin{figure}[!htbp]
	\centering
	\includegraphics[width=0.9\linewidth]{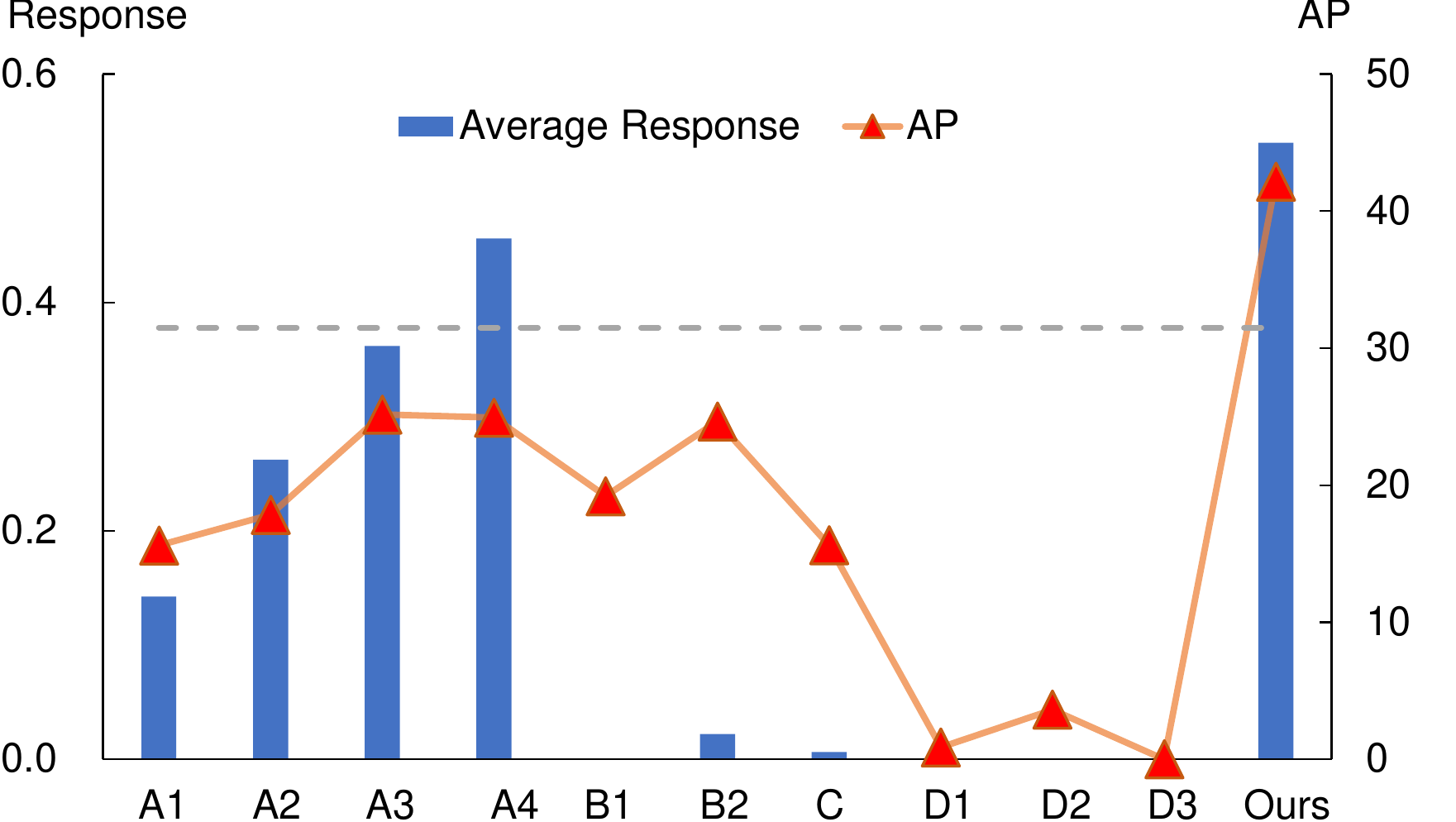}
	\caption{Results of failed attempts. A1-A4 represent approaches which use confident predictions to teach the network with confidence thresholds of $0.2$, $0.4$, $0.6$ and $0.8$, respectively. B1-B2 represent mean teachers with EMA parameters of $0.99$ and $0.999$. C represents the re-weighting method. D1-D3 represents easy-easy, hard-hard and hard-easy augmentation strategies. The gray dash lines is the AP of the initial supervised model.
}
	\label{fig:attempt}
\end{figure}

We present some failed attempts to avoid collapsing.  The first is to balance foreground/background pixels by class re-weighting. Since we do not have labels, we assign larger weights for pixels with larger heatmap predictions since they are more likely from foreground. We tried several weight functions but collapsing still occurs because we do not have ground-truth labels (see Figure \ref{fig:attempt} (C)). The second approach uses confident predictions to teach the network. If the maximum response of a heatmap (of teacher) is larger than a threshold, we use it as supervision. Otherwise, we do not use it in loss computation. When the threshold is small, the performance is much worse than the initial supervised model (see (A1)). When the threshold is large, very few pixels are involved in training and the performance is barely improved (see (A4)). The third approach stabilizes training using Mean Teacher \cite{tarvainen2017mean} in which the teacher is the exponential moving average of the student $\theta' \leftarrow \alpha \theta' + (1-\alpha) \theta$. We set $\alpha$ to be $0.99$ and $0.999$, respectively. The performance is worse than the initial supervised model which does not use unlabeled images (see (B1-B2)). 

\begin{table*}[!tbp]
	\centering
	\caption{Comparison to the state-of-the-art methods on the COCO \emph{TEST-DEV} dataset. The COCO \emph{Train} set is the labeled set and COCO \emph{WILD} set is the unlabeled set. The person detection results are provided by Simple Baseline \cite{simplebaselines} and flipping strategy is used. }
	\label{table:cocotest}
	\small
\begin{tabular}{llccclccccc}
\toprule
\textbf{Method} & \textbf{Network} & \multicolumn{1}{l}{\textbf{Input Size}} & \multicolumn{1}{l}{\textbf{GFLOPS}} & \multicolumn{1}{l}{\textbf{\#Params}} & \multicolumn{1}{l}{\textbf{AP}} & \multicolumn{1}{l}{\textbf{AP0.50}} & \multicolumn{1}{l}{\textbf{AP0.75}} & \multicolumn{1}{l}{\textbf{APM}} & \multicolumn{1}{l}{\textbf{APL}} & \multicolumn{1}{l}{\textbf{AR}} \\
    \hline
SB \cite{simplebaselines}  & ResNet50            & 256 × 192                               & 8.9                                 & 34.0                                    & 70.2                  & 90.9                                & 78.3                                & 67.1                             & 75.9                             & 75.8                            \\
SB \cite{simplebaselines}  & ResNet152           & 256 × 192                               & 15.7                                & 68.6                                    & 71.9                & 91.4                                & 80.1                                & 68.9                             & 77.4                             & 77.5                            \\
HRNet \cite{sun2019hrnet}          & HRNetW48         & 384 × 288                               & 32.9                                & 63.6                                    & 75.5               & 92.5                                & 83.3                                & 71.9                             & 81.5                             & 80.5                            \\

MSPN  \cite{li2019rethinking}          & ResNet50       & 384 × 288                               & 58.7                                & 71.9                                    & 76.1                 & 93.4                                & 83.8                                & 72.3                             & 81.5                             & 81.6                            \\
DARK \cite{zhang2020distribution}  & HRNetW48         & 384 × 288                               & 32.9                                & 63.6                                    & 76.2                & 92.5                                & 83.6                                & 72.5                             & 82.4                             & 81.1                            \\
UDP \cite{huang2020devil}  & HRNetW48         & 384 × 288                               & 33.0                                & 63.8                                    & 76.5                & 92.7                                & 84.0                                & 73.0                             & 82.4                             & 81.6                            \\
    \hline
Ours (+SB)            & ResNet50            & 256 × 192                               & 8.9                                 & 34.0                                    & \textbf{72.3 {\footnotesize ($\uparrow$ 2.1)}}             & 91.8                                & 80.5                                & 69.3                             & 77.8                             & 77.7                            \\
Ours (+SB)            & ResNet152           & 256 × 192                               & 15.7                                & 68.6                                    & \textbf{73.7 {\footnotesize ($\uparrow$ 1.8)}}             & 92.1                                & 82.1                                & 71.0                             & 79.0                             & 79.1                            \\
Ours (+HRNet)           & HRNetW48         & 384 × 288                               & 32.9                                & 63.6                                    & \textbf{76.7 {\footnotesize ($\uparrow$ 1.2)}}             & 92.5                                & 84.3                                & 73.5                             & 82.5                             & 81.8                            \\
Ours (+DARK)       & HRNetW48         & 384 × 288                               & 32.9                                & 63.6                                    & \textbf{77.2 {\footnotesize ($\uparrow$ 1.0)}}                  & 92.6                                & 84.5                                & 73.9                             & 82.9                             & 82.2   \\ 
\bottomrule         
\end{tabular}
\end{table*}

\subsection{Performance with Many Labels}
We use COCO \emph{TRAIN} and \emph{WILD} as labeled and unlabeled datasets, respectively. The results on the \emph{VAL} set are in Table \ref{table:largescale}.  Our approach consistently outperforms the initial supervised model. It suggests that even when we have access to many labels, it still gets decent improvement with unlabeled images. We also test our approach in a more realistic setting where labeled and unlabeled images are from different datasets of MPII and AIC, respectively. Table \ref{table:mpii_test} shows the results on the test set of MPII.  Our approach outperforms all other methods. The experiment validates the values of using unlabeled images. The last two methods use extra labels and larger image sizes. 

\begin{table}[!tbp]
	\footnotesize
	\renewcommand\tabcolsep{4.0pt}
	\centering
	\caption{Comparisons on the MPII test set (PCKh@0.5). Our method uses HRNetW32 as backbone and size is  $256 \times 256$. The MPII and AIC \emph{(w/o labels)} dataset are used for training. The $*$ means extra labels in AIC are used. }
	\label{table:mpii_test}
    \begin{tabular}{lccccccccc}
    \toprule
    Method & Hea& Sho& Elb& Wri& Hip& Kne& Ank & Total \\
    \hline
    Newell et al.\cite{newell2016stacked} & 98.2  & 96.3  & 91.2  & 87.1  & 90.1  & 87.4 & 83.6 & 90.9 \\
    Xiao et al. \cite{simplebaselines}  & 98.5 & 96.6 & 91.9 & 87.6 & 91.1 & 88.1 & 84.1 & 91.5 \\
    Ke et al. \cite{ke2018multi}& 98.5  & 96.8  & 92.7  & 88.4  & 90.6  & 89.4 & 86.3 & 92.1 \\
    Sun et al. \cite{sun2019hrnet} & 98.6 & 96.9 & 92.8 & 89   & 91.5 & 89   & 85.7 & 92.3 \\
    Zhang et al. \cite{zhang2019human}& 98.6 & 97.0   & 92.8 & 88.8 & 91.7 & 89.8 & 86.6 & 92.5 \\
    \textbf{Ours}  & 98.7  & 97.3  & 93.7  & 90.2  & 92.0  & 90.3 & 86.5 & \textbf{93.0}  \\
    \hline
    Su et al.*\cite{su2019cascade}& 98.7  & 97.5  & 94.3  & 90.7  & 93.4  & 92.2 & 88.4 & 93.9  \\
    Bin et al.*\cite{bin2020adv}& 98.9 & 97.6 & 94.6 & 91.2 & 93.1 & 92.7 & 89.1 & 94.1    \\
    Bulat et al.*\cite{bulat2020toward} & 98.8 & 97.5 & 94.4 & 91.2 & 93.2 & 92.2 & 89.3 & 94.1\\
    \bottomrule  
    \end{tabular}
\end{table}

\begin{table}[!htbp]
	\small
	\centering
	\caption{Domain adaptation results measured by MPJPE (mm) on the H36M dataset. The MPII is used as labeled set and H36M is unlabeled set. No labels from H36M are used in training.   }
	\label{table:adaptation}
	\begin{tabular}{llcccc}
		\toprule
		Method      & Training Data & Shld  & Elb  & Wri  & Mean  \\
		\hline
		Supervised & MPII*         & 40.3 & 67.0 & 89.3 & 65.5 \\
		PseudoPose   & MPII*+H36M    & 39.6 & 59.2 & \textbf{76.8} & 58.5   \\
		Ours   & MPII*+H36M    & \textbf{35.8} & \textbf{56.4} & 77.6 & \textbf{56.6}   \\
		\bottomrule 
	\end{tabular}
\end{table}

Table \ref{table:cocotest} shows the results of the state-of-the-art methods on the COCO test-dev dataset. We supplement them with our approach to learn about unlabeled images from the COCO \emph{WILD} dataset. We can see that our approach consistently improves the performance although the performance of the original methods are already very high.

\subsection{Alternative Applications}
SSL can also be used for \emph{unsupervised domain adaptation} to learn about unlabeled images from a new domain. To that end, we evaluate different methods by trying to adapt the model learned on the MPII dataset to the H36M dataset \cite{ionescu2014human3}. We first estimate $2$D poses from different camera views and then recover the $3$D pose by triangulation \cite{hartley2003multiple}.
The results are shown in Table \ref{table:adaptation}. Directly using the model trained on the MPII dataset gets a larger error. Our approach decreases the error by about $15\%$. The improvement on challenging joints such as ``elbow'' and ``wrist'' is even larger. The approach achieves better results than other SSL methods which use unlabeled images.

\begin{table}[!tbp]
	\small
	\centering
	\caption{Effect of pre-trained models on $2$D pose estimation tasks on the H36M dataset. }
	\label{table:pretrain2d}
    \begin{tabular}{lccccc}
    \toprule
    Pre-train Method       & Knee & Ankle  & Elbow  & Wrist  & Avg  \\
    \hline
    Supervised    & 92.5 & 88.8 & 88.2 & 83.3 & 88.2 \\
    PseudoPose   & 92.1 & 88.5 & 89.3 & 84.1 & 88.5 \\
    Ours        & 93.5 & 90.6 & 89.9 & 84.9 & \textbf{89.7} \\
    \bottomrule  
    \end{tabular}
\end{table}

\begin{table}[!tbp]
	\small
	\centering
	\caption{Effect of pre-trained models on $3$D pose estimation tasks on the H36M dataset. The errors are measured by MPJPE (mm) .  }
	\label{table:pretrain3d}
    \begin{tabular}{lccccc}
    \toprule
    Pre-train Method       & Knee & Ankle  & Elbow  & Wrist  & Avg  \\
    \hline
    Supervised & 38.2 & 58.0 & 39.7 & 56.2 & 48.0   \\
    PseudoPose &37.5 & 59.0 & 39.4 & 54.8 & 47.7   \\
    Ours       & \textbf{35.3} & \textbf{49.3} & \textbf{36.2} & \textbf{50.1} & \textbf{42.7}  \\
    \bottomrule  
    \end{tabular}
\end{table}

SSL can also be used for learning \emph{pre-trained models} using unlabeled images which can then be finetuned on a new dataset in a supervised way. sIn our experiment, We pre-train a $2$D pose estimator on the MPII dataset and the AIC dataset (w/o labels) using our approach and finetune it on the H36M dataset. The $2$D and $3$D pose estimation results on H36M are shown in Table \ref{table:pretrain2d} and \ref{table:pretrain3d}, respectively. We can see that the pre-trained model learned by our approach achieves notably higher $2$D pose estimation accuracy and lower $3$D pose error than the model pre-trained only on the labeled dataset MPII and finetuned on H36M.

\section{Conclusion}
In this work,  we present the first systematic study of semi-supervised $2$D pose estimation. In particular, we first identify and discuss the collapsing problem in consistency based methods. Then we present a simple yet effective approach to solve the problem. We conduct extensive experiments to validate the effectiveness of our approach and show that it can benefit many different application scenarios. We released our code and models hoping to inspire more research in this direction.

\paragraph{Acknowledgement}
This work was supported in part by MOST-2018AAA0102004, NSFC-61625201, 61527804 and DFG TRR169 / NSFC Major International Collaboration Project "Crossmodal Learning".

{\small
\bibliographystyle{ieee_fullname}
\bibliography{egbib}
}


\end{document}